# Numerical representations of acceptance

**Didier Dubois – Henri Prade**
Institut de Recherche en Informatique de Toulouse – C.N.R.S.
Université Paul Sabatier – Bât. 1R3, 118 route de Narbonne
31062 TOULOUSE Cedex – FRANCE
{dubois, prade}@irit.irit.fr

## Abstract

Accepting a proposition means that our confidence in this proposition is strictly greater than the confidence in its negation. This paper investigates the subclass of uncertainty measures, expressing confidence, that capture the idea of acceptance, what we call acceptance functions. Due to the monotonicity property of confidence measures, the acceptance of a proposition entails the acceptance of any of its logical consequences. In agreement with the idea that a belief set (in the sense of Gärdenfors) must be closed under logical consequence, it is also required that the separate acceptance of two propositions entail the acceptance of their conjunction. Necessity (and possibility) measures agree with this view of acceptance while probability and belief functions generally do not. General properties of acceptance functions are established. The motivation behind this work is the investigation of a setting for belief revision more general than the one proposed by Alchourrón, Gärdenfors and Makinson, in connection with the notion of conditioning.

## 1 INTRODUCTION

Gärdenfors (1988, 1992) views a belief state as modelled by a so-called *belief set* made of logical sentences. The key property of a belief set is its closure under logical consequences. Namely if a belief set $\mathcal{K}$ logically entails a formula $\phi$, $\phi$ also belongs to $\mathcal{K}$. Alchourrón, Gärdenfors and Makinson (1985) have proposed rationality postulates (the so-called AGM postulates) for belief revision. Gärdenfors and Makinson (1988) have then established that revision operations obeying these postulates are based on an epistemic entrenchment ordering. Such an ordering has been shown to be equivalent to a comparative necessity relation (Dubois and Prade, 1991) whose unique numerical counterparts are set functions called necessity measures (Dubois, 1986). Thus, a set function modelling (un)certainty is at work in the belief revision operations obeying the AGM postulates.

A natural question then arises: can we imagine other revision mechanisms underlied by other types of set functions? A preliminary investigation of this problem is addressed in this paper in the following way. A belief state will be directly described by means of a set function, expressing confidence, which is assumed to be monotonic with respect to set inclusion, as usual for any set function candidate for modelling uncertainty. A belief base is then made of propositions $\phi$ (here viewed as sets of models) which are accepted in the sense that the confidence in $\phi$ is strictly greater than the confidence in "not $\phi$". In the paper we shall assume that this belief base is a belief set, that is, closed under logical consequences. Namely, if $\phi$ entails $\psi$ and $\phi$ is accepted, $\psi$ should be accepted (a condition satisfied by any confidence function due to their monotonicity), and moreover, if $\phi$ is accepted and $\psi$ is accepted, then the conjunction $\phi \wedge \psi$ should be accepted (closure under conjunction). The latter induces constraints on the set functions which can be used for defining a belief set in this way. Confidence measures producing a belief base that is closed under conjunction are called acceptance functions. Section 2 formally introduces the proposed view of belief sets in terms of confidence measures. Section 3 investigates the characteristic properties of acceptance functions. Section 4 provides various examples of acceptance functions (including necessity measures) and characterizes the probability measures as well as the belief functions in the sense of Shafer, that are acceptance functions. Section 5 defines conditioning for acceptance functions and briefly discusses its use for belief revision in this setting.

## 2 ACCEPTED BELIEFS

Let $\Omega$ be a *finite* set of states representing possible situations. A subset of $\Omega$ can be understood as the set of models of a proposition, also called an event. Capital letters A, B, C,... will both denote propositions and their corresponding sets of models. A [0,1]-valued *confidence measure* g on $2^\Omega$, also called a "fuzzy" measure (Sugeno, 1977), is a monotonic, [0,1]-valued, set function such that

i) $g(\emptyset) = 0$ ; ii) $g(\Omega) = 1$;
iii) if $A \subseteq B \subseteq \Omega$, then $g(A) \leq g(B)$.

i) and ii) mean respectively that no confidence can be put

on the contradiction, and that the tautology should receive the maximum amount of confidence, while iii) expresses the compatibility of the confidence measure with the logical entailment: if A entails B, the confidence in B cannot be less than the confidence in A. Besides, the use of the scale [0,1] is not compulsory for defining confidence measures. A linear ordering is sufficient. However the use of the scale [0,1] is natural if we want to include classical uncertainty set functions such as probability measures or belief functions.

In the literature, set functions representing uncertainty, here called confidence measures, are always supposed to be monotonic with respect to set inclusion. Thus confidence measures encompass probability measures, belief and plausibility functions, possibility and necessity measures among other well-known uncertainty functions.

*Acceptance* is understood as taking a proposition for granted, as being more likely than its negation. Formally it is then defined by

$$A \text{ is accepted} \quad \text{iff} \quad g(A) > g(\bar{A})$$

where $\bar{A}$ denotes the complement/negation of A. This concept of acceptance is in accordance with the view that it is not really a matter of degree, as advocated by Cohen (1993), because accepting A comes down to a mental decision, namely "to adopt the policy of taking the proposition that A as a premiss in certain circumstances". A belief base is then made of all the accepted propositions in a given context. Namely let $\mathcal{AC}(g) = \{A, g(A) > g(\bar{A})\}$ be the belief base induced by g. Propositions out of the belief base are either such that the opposite proposition is in the belief base or such that an equal confidence is given to the proposition and its opposite. In the latter case, it expresses a state of ignorance with respect to the proposition. The assumed closure of the belief base leads to postulate that

1) if A is accepted and A implies B then B is accepted;
2) if A is accepted and B is accepted, then $A \cap B$ is accepted.

When these postulates hold, $\mathcal{AC}(g)$ is a belief set in the sense of Gärdenfors. These two conditions write

i) if $g(A) > g(\bar{A})$ and $A \subseteq B$ then $g(B) > g(\bar{B})$,

this is satisfied by any confidence measure since

$$g(B) \geq g(A) > g(\bar{A}) \geq g(\bar{B}).$$

Indeed A implies B is expressed by $A \subseteq B$ or equivalently to $\bar{A} \supseteq \bar{B}$.

ii) $\left.\begin{array}{l} g(A) > g(\bar{A}) \\ g(B) > g(\bar{B}) \end{array}\right\} \Rightarrow g(A \cap B) > g(\bar{A} \cup \bar{B})$.

(closure under conjunction)

This second requirement is not innocuous at all, but has important consequences on the type of confidence measures which can be used as it is shown in the next section. Confidence measures satisfying the two requirements are called *acceptance functions* in the following. Note that requirement ii) forbids to have A accepted and B accepted with $A \cap B = \emptyset$ since $g(\emptyset) = 0$ should hold for confidence measures, which leads to the violation of ii). Thus the set of accepted propositions $\{A, g(A) > g(\bar{A})\}$ is deductively closed and consistent when g is an acceptance function.

## 3 GENERAL PROPERTIES OF ACCEPTANCE FUNCTIONS

Acceptance functions are characterized by the existence of a kernel subset K defined by the intersection of all the accepted subsets, such that any accepted subset/proposition A can be determined by checking that $A \supseteq K$. This characterization is made precise by the different propositions established in this section.

**Proposition 1:** Let g be an acceptance function. Then $\exists! K \subseteq \Omega, K \neq \emptyset$, such that $g(K) > g(\bar{K})$
and    i)   $\forall A \supseteq K$, A is accepted;
       ii)  $\forall A \subseteq \bar{K}$, $\bar{A}$ is accepted;
       iii) $\forall A$ such that $A \not\supseteq K$ and $A \not\subseteq \bar{K}$, $g(A) = g(\bar{A})$.

Proof:
> Let g be an acceptance function.
> Let $K = \bigcap \{A_i \mid g(A_i) > g(\bar{A_i})\}$.
> Note that if $A_i$ is accepted and $A_j$ is accepted, then $A_i \cap A_j$ is accepted; moreover if $A_k$ is also accepted, then $(A_i \cap A_j) \cap A_k$ is accepted and so on. Thus $\bigcap \{A_i \mid g(A_i) > g(\bar{A_i})\}$ is accepted and denoted K. As such, it cannot be empty (since $g(\emptyset) = 0$ for confidence measures). Then $K \neq \emptyset$ and $g(K) > g(\bar{K})$.
> Due to the monotonicity of g we have
> $g(A) \geq g(K)$ and $g(\bar{K}) \geq g(\bar{A})$, $\forall A, A \supseteq K (\Leftrightarrow \bar{A} \subseteq \bar{K})$
> and
> $g(\bar{A}) \geq g(K) > g(\bar{K}) \geq g(A)$, $\forall A, A \subseteq \bar{K} (\Leftrightarrow \bar{A} \supseteq K)$.
> Assume $A \not\supseteq K$ and $A \not\subseteq \bar{K}$, or equivalently $\bar{A} \cap K \neq \emptyset$ and $A \cap K \neq \emptyset$.
> Then if A is accepted, $A \cap K$ should be accepted since K is accepted. This contradicts the fact that K is the smallest accepted subset. Thus $g(A) \leq g(\bar{A})$. But if $\bar{A}$ is accepted, $\bar{A} \cap K$ should be accepted which again leads to a contradiction. Thus $g(\bar{A}) \leq g(A)$ and finally $g(A) = g(\bar{A})$.    ∎

Figure 1 visualizes the contents of the above proposition.

An immediate consequence of Proposition 1 is that

A is accepted if and only if $A \supseteq K$
(since $A \subseteq \bar{K} \Leftrightarrow \bar{A} \supseteq K$ in case ii)).

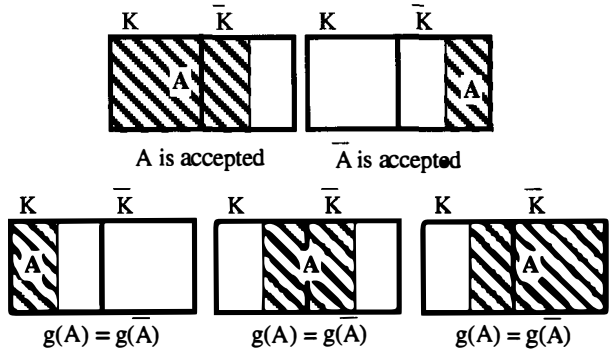

Figure 1

Moreover it can be shown that all the subsets A such that neither A nor $\bar{A}$ lies in $\mathcal{AC}(g)$, have the same level of confidence. In other words, the following proposition holds.

**Proposition 2:** Let g be an acceptance function. $\exists k \in [0,1]$, $\forall A$ such that $A \not\supseteq K$ and $A \not\subseteq \bar{K}$, $g(A) = g(\bar{A}) = k$. Moreover $g(\bar{K}) \leq k$ and $k \leq g(K)$ holds.

Proof:
> Assume that
> $\exists A, \bar{A} \cap K \neq \emptyset, A \cap K \neq \emptyset, g(A) = g(\bar{A}) = k$
> $\exists B, \bar{B} \cap K \neq \emptyset, B \cap K \neq \emptyset, g(B) = g(\bar{B}) = k'$
> with $k > k'$.
> Clearly, $(A \cup B) \cap K \neq \emptyset$. Let us show that $g(A \cup B) \neq g(\bar{A} \cap \bar{B})$. Indeed
> $g(A \cup B) \geq g(A) = k > k' = g(\bar{B}) \geq g(\bar{A} \cap \bar{B})$.
> Then we have $\overline{(A \cup B)} \cap K = \emptyset$ or equivalently, $(A \cup B) \supseteq K$.
> We can prove in the same way that
> $(\bar{A} \cup B) \supseteq K$
> $(A \cup \bar{B}) \supseteq K$
> $(\bar{A} \cup \bar{B}) \supseteq K$
> since A and $\bar{A}$, B and $\bar{B}$ play the same role. Then
> $K \subseteq (A \cup B) \cap (\bar{A} \cup B) \cap (A \cup \bar{B}) \cap (\bar{A} \cup \bar{B}) = [(A \cap \bar{A}) \cup B] \cap [(A \cap \bar{A}) \cup \bar{B}] = \emptyset$.
> This is impossible and leads to the rejection of the hypothesis $k > k'$. Similarly, $k' > k$ should be also rejected. Hence $g(A) = g(B) = k = k'$. When $A \neq \emptyset$, $A \subseteq K$ and $A \neq K$, note that $\bar{A}$ verifies the assumptions at the beginning of the proof, i.e., $\bar{A} \cap K \neq \emptyset$, $A \cap K \neq \emptyset$. Thus $g(A) = g(\bar{A}) = k$ since $A \notin \mathcal{AC}(g)$ and $\bar{A} \notin \mathcal{AC}(g)$. On the whole, $\exists k$, $\forall A \notin \mathcal{AC}(g), \bar{A} \notin \mathcal{AC}(g), g(A) = k$. Choosing $A \subseteq K$, then if $A \neq \emptyset$
> $g(A) = k \leq g(K)$.
> Similarly for $A \supseteq \bar{K}, A \neq \Omega$, we have
> $g(\bar{K}) \leq g(A) = k$. ∎

Thus we have shown that for any acceptance function g there exists a kernel subset $K \subseteq \Omega$, and a number $k \in [0,1]$ such that $g(K) > g(\bar{K})$, $k \in [g(\bar{K}), g(K)]$, and g verifies

$$\begin{cases} g(\Omega) = 1 \\ \text{if } K \subseteq A \text{ then } g(A) \geq g(K) \\ \text{if } K \subseteq \bar{A} \text{ then } g(\bar{K}) \geq g(A) \\ \text{if } K \not\subseteq A \text{ and } K \not\subseteq \bar{A} \text{ then } g(A) = k \in [g(\bar{K}), g(K)] \\ g(\emptyset) = 0. \end{cases} \quad (1)$$

k will be called the *indifference level* of g. Conversely, we have

**Proposition 3:** Any confidence measure satisfying (1) is an acceptance function.

Proof:
> Let A be such that $g(A) > g(\bar{A})$. Then $K \subseteq A$. Let B be another accepted subset. Then $K \subseteq B$ and $K \subseteq A \cap B$ with $g(A \cap B) \geq g(K) > g(\bar{K}) \geq g(\bar{A} \cup \bar{B})$. ∎

Thus the propositions such that A or its negation is entailed by K can be ordered according to the values of $g(A)$ or $g(\bar{A})$, while the propositions such that neither themselves nor their opposites are accepted, are all put by g at the *same level k*, expressing mere ignorance about them. In particular, we find that any acceptance function g is also a partial ignorance function in the sense of Dubois, Prade and Smets (1995) that maps all unknown propositions to some level k between 0 and 1.

Acceptance is preserved by duality as shown by the following proposition.

**Proposition 4:** Let h be the set function associated with g by duality. Namely $\forall A, h(A) = 1 - g(\bar{A})$. Then h is an acceptance function if and only if g is an acceptance function, and h and g have the same kernel subset and the sum of the indifference levels of h and g equals 1.

Proof:
> Let g be an acceptance function with kernel subset K and indifference level k. Then from Propositions 1 and 2, we have $h(\Omega) = 1, h(\emptyset) = 0$; when $K \subseteq A$, we have $g(A) \geq g(K) > g(\bar{K}) \geq g(\bar{A}) \Leftrightarrow h(A) \geq h(K) > h(\bar{K}) \geq h(\bar{A})$ and when $K \not\subseteq A$ and $K \not\subseteq \bar{A}$, $g(A) = g(\bar{A}) = k \Leftrightarrow h(A) = h(\bar{A}) = 1 - k$. Thus applying Proposition 3, if g is an acceptance function then h is an acceptance function (with the same kernel subset K and indifference level $1 - k$). The converse obviously holds since $\forall A, g(A) = 1 - h(\bar{A})$. ∎

## 4 SET FUNCTIONS DEFINING A BELIEF SET

### 4.1 EXAMPLES OF ACCEPTANCE FUNCTIONS

Let us consider a belief function in the sense of Shafer (1976) defined by $Bel(A) = \Sigma_{\emptyset \neq C \subseteq A} m(C)$ where m is a basic probability assignment such that $\Sigma_C m(C) = 1$; C is called a focal subset iff $m(C) > 0$. Belief functions are monotonic w.r.t. set inclusion. Note that the expression of acceptance for A in terms of belief functions, namely Bel(A) > Bel($\bar{A}$) is equivalent to its expression in terms of plausibility measures Pl(A) = 1 − Bel($\bar{A}$), since the equivalence Bel(A) > Bel($\bar{A}$) ⇔ Pl(A) > Pl($\bar{A}$) obviously holds.

The closure under conjunction writes

$$\forall A, \forall B, \begin{cases} \Sigma_{\emptyset \neq C \subseteq A} m(C) > \Sigma_{\emptyset \neq C \subseteq \bar{A}} m(C) \\ \Sigma_{\emptyset \neq C \subseteq B} m(C) > \Sigma_{\emptyset \neq C \subseteq \bar{B}} m(C) \end{cases}$$
$$\Rightarrow \Sigma_{\emptyset \neq C \subseteq A \cap B} m(C) > \Sigma_{\emptyset \neq C \subseteq \bar{A} \cup \bar{B}} m(C) \quad (2)$$

Let us consider a belief function whose focal subsets are structured in the following way: $\exists K$, m(K) > 0, and $\forall C$ if m(C) > 0 then K $\subseteq$ C. K is called a *core* subset and the intersection of all the focal subsets contains the core focal subset. Then we have the following result.

**Proposition 5:** A belief function with a core subset is an acceptance measure.

Proof:
> Indeed K $\subseteq$ A and K $\subseteq$ B are necessary conditions to have Bel(A) > 0 and Bel(B) > 0. Then K $\subseteq$ A $\cap$ B and Bel(A $\cap$ B) > 0. Since $\forall C$ such that m(C) > 0, C $\supseteq$ K, we have A $\cap$ C $\neq \emptyset$, B $\cap$ C $\neq \emptyset$, A $\cap$ B $\cap$ C $\neq \emptyset$, then Bel($\bar{A}$) = Bel($\bar{B}$) = Bel($\bar{A} \cup \bar{B}$) = 0 and (2) holds trivially. ∎

Clearly A is accepted if and only if A $\supseteq$ K and for the propositions A which are accepted we have Bel(A) $\geq$ Bel(K) > Bel($\bar{K}$) = 0. Moreover, k, as defined by Proposition 2, is equal to 0, i.e., k = Bel($\bar{K}$) = 0. In terms of plausibility measures, we have Pl(A) = Pl(K) = 1 > Pl($\bar{K}$) = 1 − Bel(K) > 0 if and only if A is accepted and k = Pl($\bar{K}$) = 1.

Let us consider necessity and possibility measures, denoted by N and $\Pi$ respectively (Zadeh, 1978, Dubois and Prade, 1988). They are defined from a ranking of the elements of $\Omega$ according to a possibility distribution $\pi$: $\Omega \to [0,1]$, such that $\max_\omega \pi(\omega) = 1$ and $\pi(\omega) \geq \pi(\omega')$ means that $\omega$ is at least as plausible as $\omega'$. The set functions are then defined as $\Pi(A) = \max_{\omega \in A} \pi(\omega)$ and $N(A) = \min_{\omega \notin A} 1 − \pi(\omega)$.

**Corollary:** Necessity measures (as well as possibility measures) are acceptance measures.

Proof:
> Indeed necessity measures (e.g., Dubois and Prade, 1988) are consonant belief functions whose focal subsets are nested, and the smallest focal subset is the kernel. This is not surprizing since we have both N(A) > 0 $\Rightarrow$ N($\bar{A}$) = 0 and N(A $\cap$ B) = min(N(A), N(B)), while possibility measures (Zadeh, 1978) are associated with necessity measures by the duality $\Pi(A) = 1 − N(\bar{A})$, and then $\Pi(A) > \Pi(\bar{A}) \Leftrightarrow N(A) \geq N(\bar{A})$ holds under the form $\Pi(A) = 1 > \Pi(\bar{A}) \Leftrightarrow N(A) > 0 = N(\bar{A})$. ∎

Let us point out that an acceptance measure *does not necessarily satisfy* the entailment g(A) > 0 and g(B) > 0 $\Rightarrow$ g(A $\cap$ B) > 0, which might look as a weak version of acceptance. Possibility measures provide a counterexample since we may have $\Pi(A) > 0$, $\Pi(B) > 0$ and $\Pi(A \cap B) = 0$ for $\Pi(A) = \sup_{\omega \in A} \pi(\omega)$, with $\pi > 0$ on A $\cap \bar{B}$ and $\bar{A} \cap$ B while $\pi = 0$ on A $\cap$ B. Besides, note that necessity measures do not only satisfy the closure under conjunction under the form

$$\begin{cases} N(A) > N(\bar{A}) = 0 \\ N(B) > N(\bar{B}) = 0 \end{cases} \Rightarrow N(A \cap B) > N(\bar{A} \cup \bar{B}) = 0$$

but also fulfil the property

$$\begin{cases} N(A) = 1 \\ N(B) = 1 \end{cases} \Rightarrow N(A \cap B) = 1.$$

More generally, let us consider an acceptance function such that g(A) = 1 and g(B) = 1 entails g(A $\cap$ B) = 1 for all A and B. This means that $\mathcal{AC}(g)$ contains {A, g(A) = 1}, which is a sub-belief set of $\mathcal{AC}(g)$. Indeed, it cannot be that g(A) = g($\bar{A}$) = 1 (which would imply g($\emptyset$) = 1). Then $\exists K^*$ such that g(A) = 1 if and only if A $\supseteq K^*$ with $K^* = \bigcap \{A \mid g(A) = 1\}$. Clearly $K^* \supseteq K = \bigcap \{A \mid g(A) > g(\bar{A})\} = \bigcap \{A \mid N(A) > 0\}$ when g = N. Since N(A) = $\inf_{\omega \notin A} (1 − \pi(\omega))$, $K^* = \{\omega \in \Omega, \pi(\omega) > 0\}$ and K = $\{\omega \in \Omega, \pi(\omega) = 1\}$.

Besides, there exist belief functions without a core subset, which are acceptance measures.

**Proposition 6:** A belief function with a focal subset which is a singleton and has a weight strictly greater than one half, is an acceptance measure.

Proof:
> Let $\{\omega_0\}$ be the (necessarily unique) focal singleton of a belief function Bel such that m($\{\omega_0\}$) > 1/2. If A is accepted Bel(A) > Bel($\bar{A}$) which entails that $\omega_0 \in$ A. Indeed assume $\omega_0 \notin$ A, then $\omega_0 \in \bar{A}$ and Bel($\bar{A}$) > 1/2, then Bel(A) < 1/2 since Bel(A) + Bel($\bar{A}$) < 1; this leads to a contradiction. Thus if A is accepted and B is accepted, $\omega_0$ belongs to both A and B and thus belongs to A $\cap$ B. Then A $\cap$ B is accepted, since Bel(A $\cap$ B) > 1/2 > Bel($\bar{A} \cup \bar{B}$). ∎

The state $\omega_0$ is understood as the "normal" one in the sense that this state is more believed to be the true one than any other statement that excludes it. Note that apart from this focal subset $\{\omega_0\}$, there is no constraints on the other focal subset(s) (which exist if m($\{\omega_0\}$) < 1), except that the sum of their weights is 1 − m($\{\omega_0\}$) $\leq$ 1/2.

Observe that with a belief function whose focal subsets have a non-empty intersection, but without a core subset, we may have Bel(A) > Bel($\bar{A}$), Bel(B) > Bel($\bar{B}$) while Bel(A ∩ B) = 0, since the intersection A ∩ B may no longer contain a focal subset. Indeed consider, for instance, a belief function whose all focal subsets $F_i$ have a non-empty intersection, i.e., $\bigcap_i F_i \neq \emptyset$, but the intersection is not a focal subset. Then Bel($F_i$) > 0 = Bel($\overline{F_i}$), $\forall i$. Assume moreover that the belief function is such that $\exists i,j$, $F_i \cap F_j$ contains no focal subset, then Bel($F_i \cap F_j$) = 0. Thus *there exist belief functions which are not acceptance functions*. In fact, a more interesting question is to investigate if there are many belief functions that can act as acceptance functions. This is the topic of the following subsection, which fully solves the problem.

## 4.2 BELIEF FUNCTIONS WHICH ARE ACCEPTANCE FUNCTIONS

The following proposition characterizes the ones and only belief functions which are acceptance functions.

**Proposition 7:** A belief function Bel (with basic probability assignment m) is an acceptance function based on a kernel K = $\bigcap${A, Bel(A) > Bel($\bar{A}$)} if and only if:
i) either there exists a singleton K such that m(K) > Bel($\bar{K}$); (then K is the kernel and |K| = 1).
ii) or any focal subset F of Bel is such that F ⊇ K where K is a focal subset such that |K| ≥ 2
iii) or the only focal subsets are {$\omega_K$}, {$\omega'_K$} with m({$\omega_K$}) = m({$\omega'_K$}), and possibly supersets of K = {$\omega_K, \omega'_K$}.

Proof:
> Assume |K| = 1. Then K = {$\omega_0$}. If m(K) > Bel($\bar{K}$), and m({$\omega_0$}) > 1/2, then Bel is an acceptance function (Proposition 6). More generally it holds that m({$\omega_0$}) > Bel($\Omega$- {$\omega_0$}) = $\sum_{\omega_0 \notin F}$ m(F). It is easy to check that $\mathcal{AC}$(Bel) is a belief set, since $\mathcal{AC}$(Bel) = {A, $\omega_0 \in$ A}. Indeed if $\omega_0 \in$ A, then Bel(A) ≥ Bel({$\omega_0$}) > Bel($\Omega$- {$\omega_0$}) ≥ Bel($\bar{A}$); and if $\omega_0 \notin$ A, then $\bar{A} \in \mathcal{AC}$(Bel), and this exhausts all possibilities. Hence Bel is an acceptance function. Conversely, if all the focal subsets F which are singletons are such that m(F) ≤ Bel($\bar{F}$), then for any singleton K, $\exists$ A ⊇ K and A is not accepted (it is sufficient to take A = K). Then Bel is not an acceptance function based on a singleton. This solves case i).
> Assume that a belief function Bel is an acceptance function based on a kernel event K with |K| ≥ 2, and such that $\exists F$, m(F) > 0 and F $\not\supseteq$ K. Then $\exists \omega_K \in$ K ∩ $\bar{F}$ and there holds Bel(F ∪ {$\omega_K$}) > Bel({$\omega_K$}). This leads to a contradiction if F ∪ {$\omega_K$} ≠ K as it is going to be shown. Indeed since Bel is supposed to be an acceptance function, we should have $\forall$A such that A $\succsim$ K, A $\not\subseteq$ $\bar{K}$, $\forall$B such that B $\succsim$ K, B $\not\subseteq$ $\bar{K}$, Bel(A) = Bel(B) from Proposition 2. But this does not hold for A = F ∪ {$\omega_K$} and B = {$\omega_K$} although A $\succsim$ K if F ∪ {$\omega_K$} $\succsim$ K, A $\not\subseteq$ $\bar{K}$ and B $\not\subseteq$ $\bar{K}$ since $\omega_K \in$ K, $\omega_K \in$ A, $\omega_K \in$ B, and B $\succsim$ K since B is a singleton and K has at least two elements. Consider now the case where F ∪ {$\omega_K$} ⊇ K but F ∪ {$\omega_K$} ≠ K. Thus, $\exists \omega \in$ F ∩ $\bar{K}$, and we have Pl({$\omega,\omega_K$}) > Pl({$\omega_K$}) since m(F) > 0. But Pl is also an acceptance function with kernel subset K from Proposition 4. Since {$\omega,\omega_K$} $\not\subseteq$ $\bar{K}$ and {$\omega,\omega_K$} $\succsim$ K (since $\omega \notin$ K and |K| ≥ 2), it should hold that Pl({$\omega,\omega_K$}) = Pl({$\omega_K$}) from Proposition 2 again, which leads to a contradiction. Hence either $\forall$F, F $\succsim$ K implies m(F) = 0, and we are in case (ii), or F ∪ {$\omega_K$} = K. Note that in case (ii) m(K) > 0 should hold; otherwise we would have Bel(K) = 0 which is impossible.
> Assume F ∪ {$\omega_K$} = K and |F| > 2. Then $\exists \omega, \omega' \neq \omega$, $\omega \in$ F, $\omega' \in$ F. Since {$\omega_K$} ⊆ K, F ⊆ K, {$\omega$} ⊆ K, we should have Bel(F) = Bel({$\omega_K$}) = Bel({$\omega$}) because Bel is supposed to be an acceptance function but Bel(F) > Bel({$\omega$}). Thus F can have only one element $\omega_F$ and we should have m({$\omega_K$}) = m({$\omega_F$}). Clearly, for any subset A disjoint with K, one must have Bel(A ∪ {$\omega_K$}) = Bel({$\omega_K$}) which forbids focal subsets disjoint with or overlapping K without containing it. This is case (iii).
> Conversely when |K| ≥ 2, Proposition 5 has established that a belief function all focal subsets of which contain a single focal subset is an acceptance function. It is easy to see that a belief function containing two singleton focal subsets of equal weight both contained in all other focal subsets is also an acceptance structure. ∎

In summary, the belief functions which are acceptance functions either have a kernel which is a focal subset contained in all other focal subsets, or have a focal singleton K of sufficient weight, or have two focal subsets which are singletons with equal weights, both included in the other focal subsets (if any).

## 4.3 PROBABILITY MEASURES WHICH ARE ACCEPTANCE FUNCTIONS

It is also interesting to characterize the probability measures which are acceptance functions. This is done by the following proposition.

**Proposition 8:** The only probability measures P that are acceptance functions are such that P({$\omega_0$}) > 1/2 for some $\omega_0$ and the probability measures such that $\exists \omega, \omega'$, P({$\omega$}) = P({$\omega'$}) = 1/2.

Proof:
> The fact that a probability measure such that $P(\{\omega_0\}) > 1/2)$ for some $\omega_0$ is an acceptance function is an immediate consequence of Proposition 6. It remains to see if they are the only ones. Let P be a probability measure such that $1/2 \geq p_1 \geq \ldots \geq p_n$ with $P(\{\omega_i\}) = p_i$. Note that $n \geq 3$, except if $p_1 = p_2 = 1/2$. Then, it is easy to see that P is an acceptance function in that latter case. Assume $n \geq 3$. Note that A is accepted if and only if $P(A) > 1/2 > P(\bar{A})$. Let $A = \{\omega_1, \ldots, \omega_i\}$ where i is the smallest rank such that $P(A) = P(\{\omega_1, \ldots, \omega_i\}) > 1/2$. Let $B = \{\omega_j, \ldots, \omega_n\}$ where j is the greatest rank such that $P(B) = P(\{\omega_j, \ldots, \omega_n\}) > 1/2$. Note, $j \leq i$, $i < n$ and $1 < j$ (since $p_1 < 1/2$). Thus $P(A \cap B) = P(\{\omega_j, \ldots, \omega_i\}) \leq 1/2$; in other words A is accepted, B is accepted but not $A \cap B$. Thus P is not an acceptance function. ∎

Probability measures which are acceptance functions such that $P(\{\omega_0\}) > 1/2)$ possess a "usual value", that is, an element which, in frequentistic terms, occurs more often than all the other together.

### 4.4 EXAMPLE OF AN ACCEPTANCE FUNCTION WHICH IS NOT A BELIEF FUNCTION

All the examples of acceptance functions which have been presented in this paper until now are belief functions. However acceptance functions which are not belief functions can be imagined. For instance, let A, B be non-empty subsets of $\Omega$, such that $A \not\subseteq B$, $B \not\subseteq A$, $A \cap B \neq \emptyset$. Let g be an acceptance function defined with $K = A \cap B$, $g(A \cap B) = k$, $g(A) = k + \varepsilon = g(B)$, $g(A \cup B) = k + 3\varepsilon$. Let C such that $B \cap C = \emptyset$ but $A \cap C \neq \emptyset$, $g(C) = \bar{k} < k$ (since $C \subseteq \bar{K}$), $g(A \cap C) = \bar{k} - 2\varepsilon$, $g(A \cup C) = k + 2\varepsilon$. Then we have

$g(A \cup B) + g(A \cap B) =$
$\qquad k + 3\varepsilon + k > g(A) + g(B) = k + \varepsilon + k + \varepsilon$
$g(A \cup C) + g(A \cap C) =$
$\qquad k + 2\varepsilon + \bar{k} - 2\varepsilon < g(A) + g(C) = k + \varepsilon + \bar{k}.$

According to Proposition 3, g is an (incompletely defined) acceptance function, but it is not a belief function since it does not hold that

$$Bel(A \cup C) + Bel(A \cap C) \geq Bel(A) + Bel(C)$$

nor a plausibility function since

$$Pl(A \cup B) + Pl(A \cap B) \leq Pl(A) + Pl(B)$$

does not hold. The above results are in agreement with the claims made by Cohen(1993) that acceptance is distinct from belief, since not only are there few belief functions and probability measures (the common representations of belief) that can act as acceptance functions but acceptance functions need not even be a subset of belief functions.

## 5 CONDITIONING ACCEPTANCE FUNCTIONS

Conditioning is the basic operation underlying revision for confidence measures. The natural way for defining the conditioning of the belief set $\{A, g(A) > g(\bar{A})\}$ where g is an acceptance function, by the subset (proposition) C is to consider the conditioned belief set of accepted beliefs $\{A \mid g(A \cap C) > g(\bar{A} \cap C)\}$, denoted by $\mathcal{AC}(g \mid C)$.

Indeed, this view of conditioning agrees both with the view of conditional statements in nonmonotonic reasoning where "if C then A generally" is understood as a constraint expressing that A is strictly more plausible than not A when C is true (e.g., Gärdenfors and Makinson, 1994), and with the way conditioning is defined for uncertainty measures like probability measures ($P(A \mid C) > 1/2 > P(\bar{A} \mid C) \Leftrightarrow P(A \cap C) > P(\bar{A} \cap C)$), or belief functions ($Pl(A \mid C) > Pl(\bar{A} \mid C) \Leftrightarrow Pl(A \cap C) > Pl(\bar{A} \cap C)$), or possibility and necessity measures ($N(A \mid C) > 0 = N(\bar{A} \mid C) \Leftrightarrow \Pi(A \cap C) > \Pi(\bar{A} \cap C)$ where $\Pi(A \mid C)$ is defined as the greatest solution of $\Pi(A \cap C) = \min(\Pi(A \mid C), \Pi(C))$ and $N(A \mid C) = 1 - \Pi(\bar{A} \mid C)$); see (Dubois and Prade, 1995) for the representation of conditional statements in terms of possibility measures.

Extending the notion of (weak) independence recently introduced for possibility measures (Dubois et al., 1994) to acceptance functions, i.e., transposing the possibilistic definition:

> A is independent from C iff $N(A) > 0$ and $N(A \mid C) > 0$

into

> A is independent from C
> iff $g(A) > g(\bar{A})$ and $g(A \mid C) > g(\bar{A} \mid C)$,

we see that an element A of a belief set still belongs to a belief set conditioned by C, if A is independent from C, which is intuitively satisfying.

This view of conditioning is clearly compatible with the "closure under entailment" property of a belief set. Namely, we have the following result:

**Proposition 9:** If g is a confidence measure, if A is such that $g(A \cap C) > g(\bar{A} \cap C)$ and $A \subseteq B$, then $g(B \cap C) > g(\bar{B} \cap C)$.

Proof:
> Let A be such that $g(A \cap C) > g(\bar{A} \cap C)$ and let B such that $A \subseteq B$, then $g(B \cap C) \geq g(A \cap C)$ since $B \supseteq A$ and $g(\bar{A} \cap C) \geq g(\bar{B} \cap C)$ since $\bar{A} \supseteq \bar{B}$ and g is a confidence measure. Consequently, $g(B \cap C) >$

$g(\bar{B} \cap C)$ and $B \in \mathcal{AC}(g \mid C)$.  ∎

Unfortunately, this view of conditioning is not generally compatible with the notion of a belief set, for all acceptance functions, because the restriction of an acceptance function to a subset C (on which the conditioning is made) is not always an acceptance function. To see it, consider, for instance a probability measure P defined by the distribution $p_1 > 1/2 > p_2 \geq ... \geq p_n$ with $P(\{\omega_i\}) = p_i$ and $C = \{\omega_2, ..., \omega_n\}$. Then we have no guarantee that $p_2 / (1 - p_1)$ is strictly greater than 1/2, and it is possible that $P(\cdot \mid C)$ is no longer an acceptance function.

The key problem here is that the "closure under conjunction" property is lost by conditioning, generally. Namely it is not true that if g is an acceptance function

$$\left.\begin{array}{l} g(S \cap C) > g(\bar{S} \cap C) \\ g(T \cap C) > g(\bar{T} \cap C) \end{array}\right\} \Rightarrow g(S \cap T \cap C) > g((\bar{S} \cup \bar{T}) \cap C) \quad (A)$$

Let us define a context-tolerant acceptance function as an acceptance function g such that for all C with $g(C) > 0$, property (A) holds. The requirement that a conditioned belief set is still a belief set means that the restriction of an acceptance function to any non-impossible subset C of $\Omega$ is still an acceptance function).

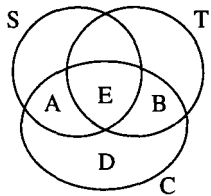

Figure 2

This property can be rewritten letting $A = S \cap C \cap \bar{T}$, $B = \bar{S} \cap C \cap T$, $E = S \cap T \cap C$, $D = C \cap \bar{S} \cap \bar{T}$ (see Figure 2):

$$\left.\begin{array}{l} g(A \cup E) > g(B \cup D) \\ g(B \cup E) > g(A \cup D) \end{array}\right\} \Rightarrow g(E) > g(A \cup B \cup D).$$

Restricting g to $A \cup B \cup E$ in the above condition (i.e., letting $D = \emptyset$) leads to a necessary condition for an acceptance function to produce a conditioned belief set that is still a belief set, namely

$\forall A, B, E$, mutually exclusive,
$$\left.\begin{array}{l} g(A \cup E) > g(B) \\ g(B \cup E) > g(A) \end{array}\right\} \Rightarrow g(E) > g(A \cup B). \quad (B)$$

It is easy to check that condition (B) implies condition (A). Indeed with the same notations as above, displayed on Figure 2, let us cut D into two disjoint subsets D' and D". Then

$g(S \cap C) > g(\bar{S} \cap C)$ and $g(T \cap C) > g(\bar{T} \cap C)$

also write

$g(A \cup E) > g(B \cup D)$ and $g(B \cup E) > g(A \cup D)$.

Hence $g(E \cup A \cup D'') > g(B \cup D')$ and $g(E \cup B \cup D') > g(A \cup D'')$. Using condition (B), one concludes $g(E) > g(A \cup B \cup D' \cup D'')$, that is $g(S \cap T \cap C) > g((\bar{S} \cup \bar{T}) \cap C)$. This establishes the equivalence between (A) and (B). Condition (B) has been independently discovered by Friedman and Halpern (1995), and characterizes context-tolerant acceptance functions.

The determination of the acceptance functions which satisfy the above condition (B), that is, such that their restriction to any subset is an acceptance function is an open problem. These acceptance functions at least clearly include possibility measures (since any possibility measure is an acceptance function and a conditioned possibility measure is still a possibility measure), but also pathological cases like acceptance functions on a 3-element set (since any set function on a 1-element or on a 2-element set is an acceptance function trivially).

We already observed that if g is an acceptance function, its dual h defined by $h(A) = 1 - g(\bar{A})$ is still an acceptance function, and that $g(A) > g(\bar{A})$ is equivalent to $h(A) > h(\bar{A})$. But the closure under conjunction conditions written in the conditioned case for g and for h restricted to C are no longer equivalent since $1 - g(\bar{A} \cap C)$ is not equal to $h(A \cap C)$ except if $C = \Omega$. This gives birth to two different ways of conditioning.

## 6 CONCLUDING REMARKS

The study of acceptance functions is relevant for the purpose of relating in a systematic way the revision of belief sets in the propositional settings to the notion of conditioning for measures of uncertainty. Indeed, it is known that Alchourrón, Gärdenfors and Makinson postulates for revision are underlied by 1) the assumption that a set of accepted beliefs is deductively closed, 2) the existence of an epistemic entrenchment relation on this belief set, which is equivalent to a necessity measure. The revision of belief sets can be expressed in terms of computing conditional possibility since any revision operation underlies a possibility measure $\Pi$ such that $A \in \mathcal{K}*_C$ if and only if $\Pi(A \cap C) > \Pi(\bar{A} \cap C)$ where $\mathcal{K}*_C$ is the belief set resulting from revising a belief set $\mathcal{K}$ by means of input C, and $N(A) = 1 - \Pi(\bar{A})$ is the numerical necessity function that represents the epistemic entrenchment relation underlying the revision of $\mathcal{K}$ (Dubois and Prade, 1992). This necessity measure exists if and only if the revision function satisfies the AGM postulates of Alchourrón, Gärdenfors and Makinson. It sounds reasonable to keep the deductive closure assumption and no longer to assume an epistemic entrenchment relation, so as to make revision in the propositional sense compatible with more theories of uncertainty.

If g is a context-tolerant acceptance function then $\mathcal{AC}(g \mid C)$ is a belief set containing C, so that g imbeds two of AGM postulates already. The next question is to figure out how many postulates are embedded in the acceptance function approach. The answer to that question will point out precisely what are the uncertainty theories that are coherent with the revision of belief sets in a logical setting.

In the acceptance function approach, it makes sense to distinguish between the cases where $C \cap K \neq \emptyset$ and $C \cap K = \emptyset$, where K is the kernel subset of the acceptance function. If $C \cap K \neq \emptyset$, conditioning on C corresponds to the idea of an expansion in the sense of Gärdenfors (1988). Note that $g(C \cap K) = k \geq g(\bar{K}) \geq g(C \cap \bar{K})$ from Proposition 2; when $g(C \cap K) > g(C \cap \bar{K})$, $C \cap K$ will play the role of the kernel subset of the acceptance measure restricted to C. When $C \cap K = \emptyset$ and $g(C) > 0$, the conditionning comes down to a genuine revision.

Another path to follow (and investigated by Friedman and Halpern (1995)) is to relate acceptance to conditional knowledge bases in the sense of Kraus, Lehmann and Magidor (1990). Namely a conditional assertion $A \mathrel{\vdash} B$ (read B is a plausible conclusion of A) is taken for granted if and only if $g(A \cap B) > g(\bar{A} \cap B)$. It is obvious that the inclusion-monotonicity of g implies that the inference relation $\mathrel{\vdash}$ satisfies the Right Weakening property ($B \subseteq B'$ and $A \mathrel{\vdash} B$ implies $A \mathrel{\vdash} B'$), that $A \mathrel{\vdash} A$ whenever $g(A) > 0$, $\forall A \neq \emptyset$. Moreover if g is a context-tolerant acceptance function then the right AND property ($A \mathrel{\vdash} B$ and $A \mathrel{\vdash} C$ implies $A \mathrel{\vdash} B \cap C$) follows. Friedman and Halpern (1995) suggest that other KLM properties are satisfied as well when property (B) holds. These results indicate that the class of useful, non-pathological acceptance functions for the purpose of belief revision (or equivalently: plausible reasoning) might not extend far beyond possibility measures.

## ACKNOWLEDGEMENTS

This work has been partially supported by the European ESPRIT Basic Research Action No. 6156 entitled "Defeasible Reasoning and Uncertainty Management Systems (DRUMS-II)".

The authors are grateful to Joe Halpern for discussions around property (B).